\documentclass{esannV2}
\usepackage[dvips]{graphicx}
\usepackage[latin1]{inputenc}
\usepackage{amssymb,amsmath,array}
\usepackage{caption}
\usepackage{multirow}
\usepackage{subfig}
\usepackage{url}
\usepackage{booktabs}
\usepackage{tikz}
\usepackage{cellspace}
\makeatletter
\DeclareRobustCommand
  \myvdots{\vbox{\baselineskip4\p@ \lineskiplimit\z@
    \hbox{.}\hbox{.}\hbox{.}}}
\makeatother

\DeclareMathOperator*{\argmax}{arg\,max}

\begin{document}
\title{Stacked Confusion Reject Plots (SCORE)}


\author{Stephan Hasler and Lydia Fischer
\vspace{.3cm}\\
%
Honda Research Institute Europe GmbH\\	
Carl-Legien-Strasse 30, 		
63073 Offenbach, Germany	
}


\maketitle

\begin{abstract}
Machine learning is more and more applied in critical application areas like health and driver assistance. To minimize the risk of wrong decisions, in such applications it is necessary to consider the certainty of a classification to reject uncertain samples. An established tool for this are reject curves that visualize the trade-off between the number of rejected samples and classification performance metrics. We argue that common reject curves are too abstract and hard to interpret by non-experts. We propose Stacked Confusion Reject Plots (SCORE) that offer a more intuitive understanding of the used data and the classifier's behavior. We present example plots on artificial Gaussian data to document the different options of SCORE and provide the code as a Python package\footnote{The source code can be accessed at \url{https://github.com/HRI-EU/score}}.
\end{abstract}

\section{Introduction}
Machine learning has entered even critical application areas because of an increased performance, but also with an increased awareness for the necessity 
for trust and explainability of its results \cite{LisboaSVFV23}.
For classification, trust can be increased by disregarding highly uncertain results \cite{chow1970optimum}.
Reject options filter classification results for certain data samples and have been realized in many settings \cite{franc2023optimal, kummert2016local, reject_survey, DBLP:conf/esann/ArteltVH22, fischer2015efficient, fischer2016optimal}.
Further, different methods were proposed to inspect the trade-off between an increasing number of rejected samples and the classification performance \cite{nadeem2009accuracy, Fischer_wsom24, condessa2017performance, pietraszek2005optimizing, hanczar2019performance}. This helps to understand the working principle of a classifier with its reject option, compare different alternatives, and to choose an optimal rejection threshold for an application.
However, the commonly used classification metrics, e.\,g.\ accuracy, recall, and, precision, mask most details about the data (like imbalanced classes) and might shadow artifacts in the classifier's behavior (like starving of a class). This makes it hard to fully understand and correctly apply results even for experts, but even more for lay people, like the user of a deployed model.
We propose Stacked Confusion Reject Plots (SCORE) as more intuitive visualization that is based on the confusions \cite{pearson1904theory} among the present classes to improve on this situation.

In Sec.~\ref{sec:related_work} we summarize related work, and present some general math and our experimental setup in Sec.~\ref{sec:methods}.
In Sec.~\ref{sec:score} we propose SCORE and showcase its variants using some artificial Gaussian data, before giving a conclusion in Sec.~\ref{sec:conclusion}.

\section{Related Work}\label{sec:related_work}
Reject options enable a classifier, which allows to define a certainty measure, to abstain from classification of single data samples. This can be implemented as a post-processing step \cite{fischer2015efficient,fischer2016optimal}. 
Besides, there are classifiers with an included reject option \cite{BakhtiariV22, villmann2016self}.
For a recent review of reject options see \cite{reject_survey, franc2023optimal}.

Often so-called global-reject options are applied, where the same rejection threshold is used across the whole input space \cite{fischer2015efficient}.
However, local reject options can be more flexible by using independent thresholds to increase reliability for specific classes or regions \cite{fischer2016optimal}.

Accuracy-Rejection Curves (ARCs) \cite{nadeem2009accuracy} are widely used to evaluate classifiers with a reject option.
They visualize the accuracy of classification with respect to the used rejection threshold and hence to the acceptance rate of the application.
Recently, \cite{Fischer_wsom24} proposed Precision-Reject Curves (PRCs) and Recall-Reject Curves (RRCs) as new alternatives for binary classification settings.
In \cite{hanczar2019performance} Cost-Reject curves are proposed that visualize the space of convex linear combinations of rejection rate and error rate.
Metrics that also consider rejected samples are proposed in \cite{condessa2017performance}. These can be used to show the effectiveness of a reject option.
Last, there is work on understanding why instances got rejected from classification \cite{DBLP:conf/esann/ArteltVH22,DBLP:conf/ijcci/ArteltBVH22}.

The mentioned metrics above have in common that they mask details of class shares and confusions that might be highly relevant to understand and apply a model. The proposed SCORE complements existing work by providing exactly those masked insights.

\section{Methods}\label{sec:methods}
Let $\mathcal{X}$ be a data set. Each sample $\boldsymbol{x} \in \mathcal{X}$ has a true class $c_{\text{t}}\in\{1, \ldots , C\}$, where $C$ is the number of classes. 
A classifier predicts a class $c_{\text{p}}$ for a sample $\boldsymbol{x}$, and additionally provides a certainty value for a defined certainty measure $r$
\begin{equation*}
    r: \mathbb{R}^n \to {\mathbb{R}^+}\,, \qquad \boldsymbol{x} \longmapsto r(\boldsymbol{x}) \,.
\end{equation*}
Let $\mathcal{X}_{\theta}\subseteq \mathcal{X}$ be the set of accepted data samples with $r(\boldsymbol{x}) \geq \theta$, where $\theta \in {\mathbb{R}^+}$ is a given threshold.
Hence the samples in $\mathcal{X}_{\theta}$ are considered reliable for the given threshold.
We focus on this so-called global reject option, where the same threshold holds for the whole input space, i.\,e., also across all classes. 

To present SCORE, we use some simple data and classifier to focus on the expressiveness of visualization instead of some specific problem and classifiers. For this, we draw class samples from a Gaussian distribution with two classes as defined in Table~\ref{tab:data}.
\begin{table}[ht]
\scriptsize
\centering
\begin{tabular}{cll}
\toprule
    $C$ & \textbf{Class} $c_{\text{t}}{=}1$  & \textbf{Class} $c_{\text{t}}{=}2$ \\
\midrule
    2 & $\mu{=}(0,0)$ $\sigma{=}(2,1)$ $n{=}40$  & $\mu{=}(2,0)$ $\sigma{=}(1,0.5)$ $n{=}100$ \\
      & $\mu{=}(4,0)$ $\sigma{=}(2,1)$ $n{=}40$ & $\mu{=}(6,0)$ $\sigma{=}(1,1)$ $n{=}60$ \\
\bottomrule
\end{tabular}
\caption{The Gaussian data used for the 2-class setting, where $\mu$, $\sigma$, and $n$ denote mean, standard deviation, and cardinality.}\label{tab:data}
\end{table}
The used classifier is an optimal Bayes, where we assume knowledge of the true data distribution. Accordingly, the predicted class $c_{\text{p}}$ of a sample and the corresponding certainty $r$ are
\begin{equation*}
    c_{\text{p}}(\boldsymbol{x}) = \argmax_{1\leq c \leq C} p(c|\boldsymbol{x})\,, \qquad r(\boldsymbol{x}) = \max_{1\leq c \leq C} p(c|\boldsymbol{x})\,.
\end{equation*}

\begin{figure}[ht]
    \centering
    \subfloat[]{{\includegraphics[width=0.49\textwidth]{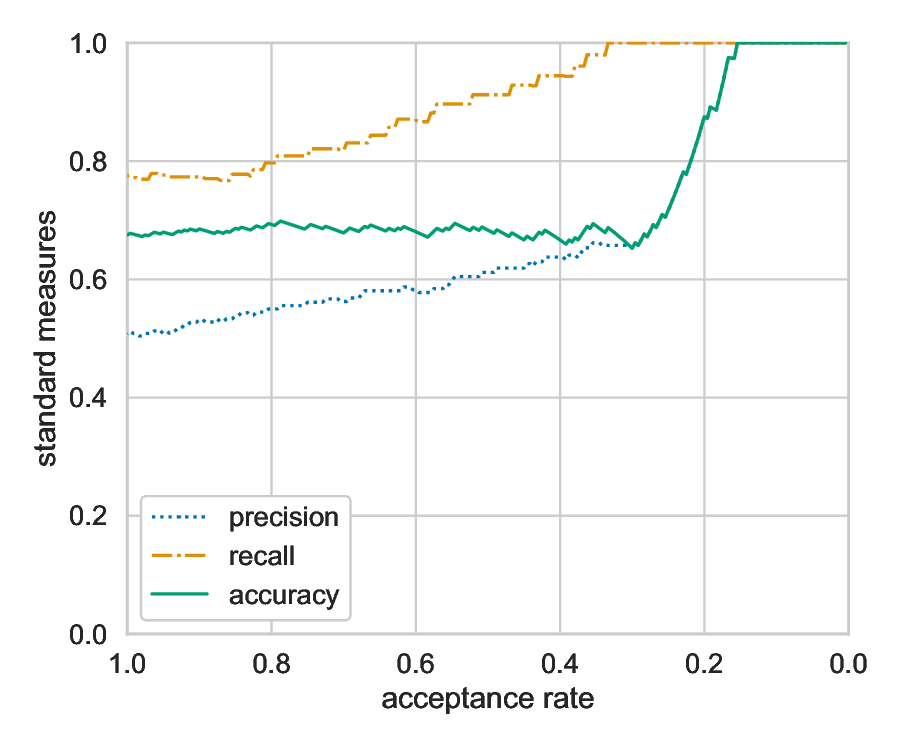}}\label{fig:example_2_classes_reject}}
    \subfloat[]{{\includegraphics[width=0.49\textwidth]{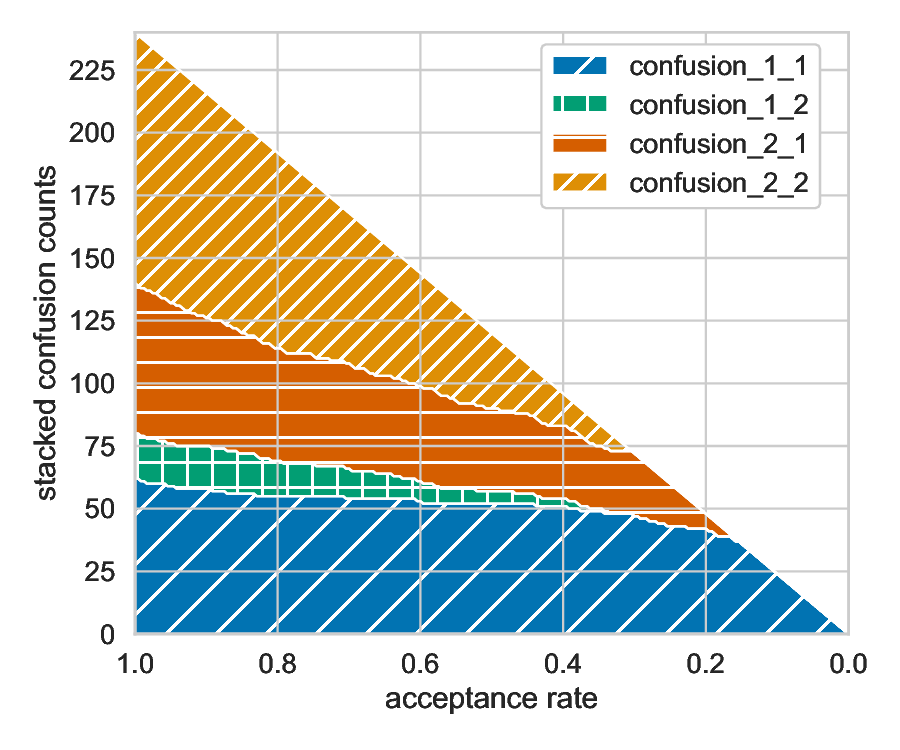}}\label{fig:example_2_classes_stack}}
\caption{Standard reject plots vs.~a variant of SCORE for the 2-class setting. a)~ACR \cite{nadeem2009accuracy}, PRC \& RRC \cite{Fischer_wsom24} might be hard to interpret. b)~Stacked confusions give more insight about the class distribution and the behavior of the classifier.}
\label{fig:standard_vs_score}
\end{figure}

A standard way of visualizing a reject option is to plot a classification evaluation metric over the acceptance rate. The acceptance rate is the ratio of $|\mathcal{X}_{\theta}|/|\mathcal{X}|$.
ARCs \cite{nadeem2009accuracy} use the accuracy as metric, which is computed on accepted samples $\mathcal{X}_{\theta}$. Likewise, PRCs and RRCs \cite{Fischer_wsom24} use precision and recall as metric, respectively. Figure~\ref{fig:example_2_classes_reject} shows these metrics for the given 2-class setting. The accuracy is rather stable for acceptance rates $> 0.3$, i.\,e., there is a fixed ratio between the number of correctly and wrongly classified samples that are rejected. But it is unclear to which classes the rejected samples are belonging to. Let us assume some medical data with $c_{\text{t}}{=}1{\hat{=}}ill$ and $c_{\text{t}}{=}2{\hat{=}}healthy$. Here it would make a difference if more \emph{ill} or \emph{healthy} persons are rejected. Also it is unclear why the accuracy strongly increases for acceptance rates $< 0.3$. 

\section{Stacked Confusion Reject Plots}\label{sec:score}
Instead of picking a specific metric as above, in SCORE we compute and visualize the full class confusion matrix for different acceptance rates.
The confusion matrix \cite{pearson1904theory} can be denoted as in Table~\ref{tab:confusion_matrix}.
\begin{table}[ht]
\scriptsize
\centering
\begin{tabular}{cc|Scccc}
  \multirow{10}{*}{\rotatebox{90}{\parbox{1.3cm}{\centering True class $c_{\text{t}}$}}} 
   & \multicolumn{5}{c}{\centering Predicted class $c_{\text{p}}$}  \\
     & & 1 & 2 & \ldots & $C$ \\
    \cline{2-6}
    & 1 & confusion\_1\_1 & confusion\_1\_2 & \ldots & confusion\_1\_$C$ \\
    & 2 & confusion\_2\_1 & confusion\_2\_2 & \ldots & confusion\_2\_$C$ \\
    & \myvdots & \myvdots & \myvdots & $\cdots$ & \myvdots \\
    & $C$ & confusion\_$C$\_1 & confusion\_$C$\_2 & \ldots & confusion\_$C$\_$C$ \\
\end{tabular}
\caption{Standard classification confusion matrix.}\label{tab:confusion_matrix}
\end{table}
For plotting we stack the elements \emph{confusion\_$c_{\text{t}}$\_$c_{\text{p}}$} of the matrix on top of each other, as shown in Fig.~\ref{fig:example_2_classes_stack}. This reveals intuitively several aspects of the data and the classifier with its reject option: There are roughly 240 samples and there is a 1-to-2 imbalance in the distribution of samples over classes. Let us interpret the data again as $c_{\text{t}}{=}1{\hat{=}}ill$ and $c_{\text{t}}{=}2{\hat{=}}healthy$.
The classifier is very correct and certain about samples of class 1, although it is the smaller class. Hence, the classifier detects \emph{ill} patients reliably among all acceptance rates $> 0.3$. Only few samples of class 1 are missed (\emph{confusion\_1\_2}). The user may decide depending on the illness if the miss rate is acceptable.
Often, the classifier confuses samples of class 2 and is more uncertain about it. Hence, \emph{healthy} people are more often classified as being \emph{ill}. Depending on the severity of the illness and the needed tests to confirm further that this prediction is true, this might lead to unacceptable consequences for the patients, e.\,g.\ psychological or physical. 
The plot supports to judge whether the risk of misclassifying a healthy person is acceptable or if a revision of the classifier, its reject option, or the anticipated acceptance rate is needed. For acceptance rates $< 0.3$, the wrong \emph{confusion\_2\_1} drops strongly, while the correct \emph{confusion\_1\_1} is rather stable which explains the strong increase in accuracy.
In this way, SCORE complements state of the art \cite{nadeem2009accuracy,hanczar2019performance,Fischer_wsom24} by offering more detailed explanations for certain aspects. 

\begin{figure}[ht]
    \centering
    \subfloat[\tiny\begin{tabular}{ll}
        type=STACK & order=CORRECT\_LAST \\
        normalize=FALSE & align=BOTTOM
    \end{tabular}]{{\includegraphics[width=0.49\textwidth]{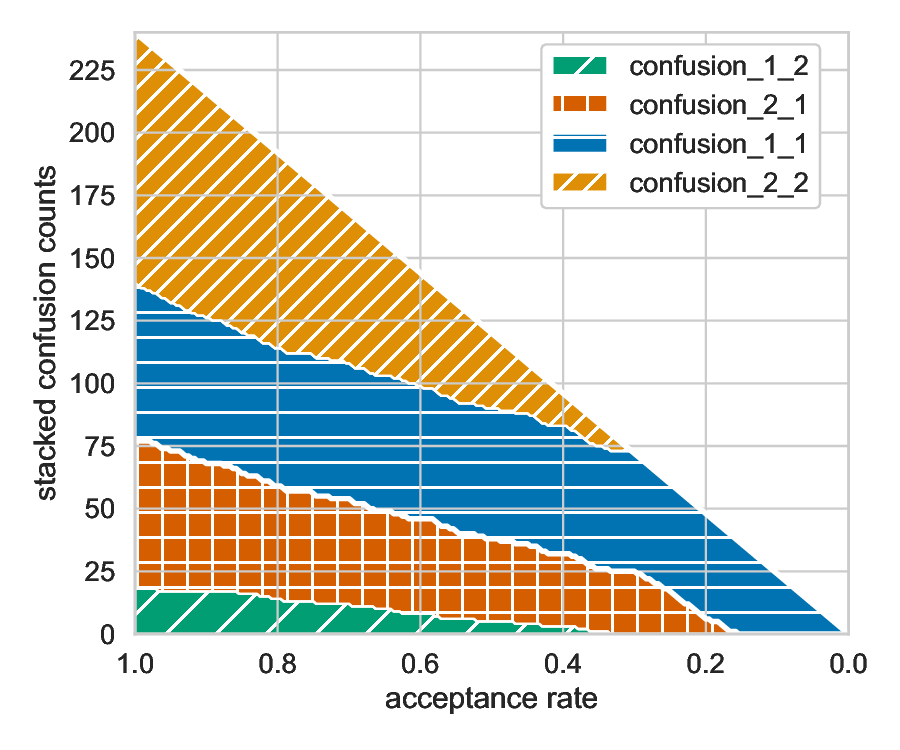}}\label{fig:example_2_classes_stack_ordered}}
    \subfloat[\tiny\begin{tabular}{ll}
        type=STACK & order=CORRECT\_LAST \\
        normalize=FALSE & align=CORRECT\_START
    \end{tabular}]{{\includegraphics[width=0.49\textwidth]{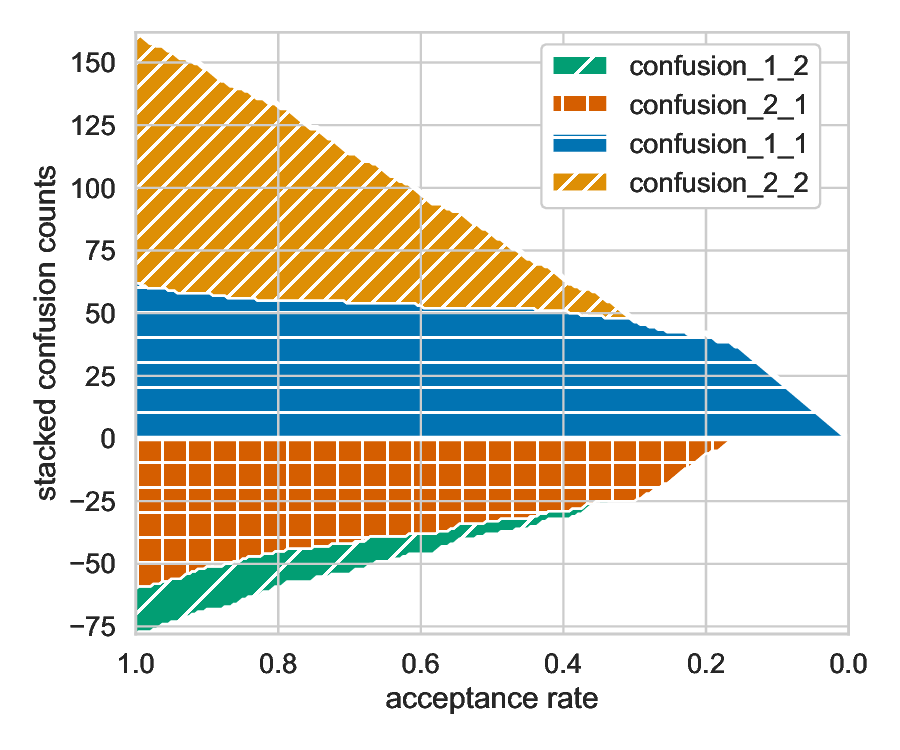}} \label{fig:example_2_classes_stack_ordered_aligned}}

    \subfloat[\tiny\begin{tabular}{ll}
        type=STACK & order=CORRECT\_LAST \\
        normalize=TRUE & align=CORRECT\_START
    \end{tabular}]{{\includegraphics[width=0.49\textwidth]{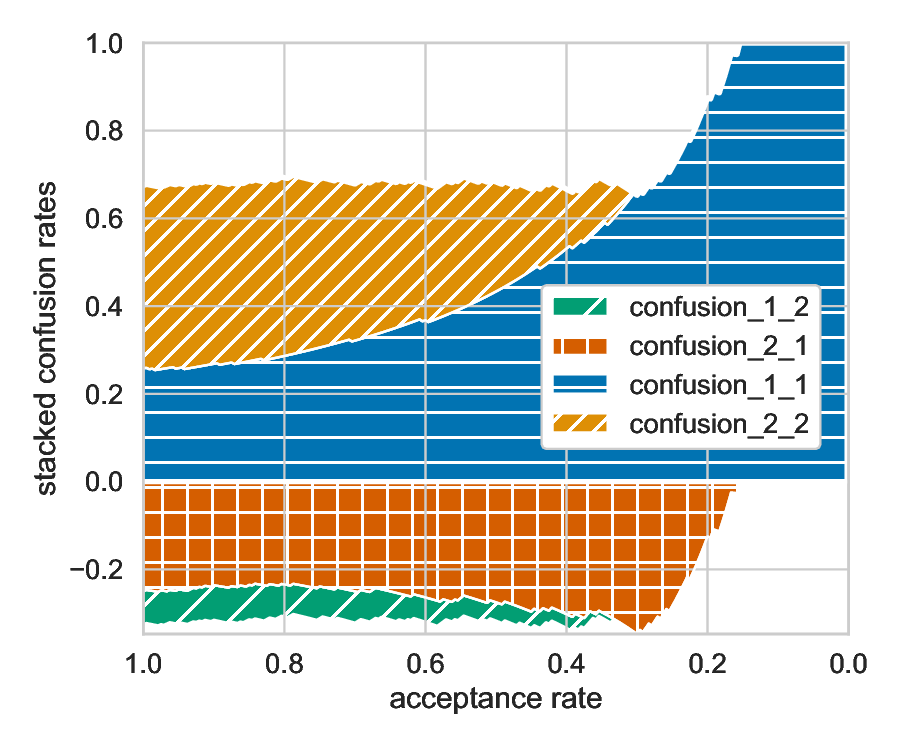}}\label{fig:example_2_classes_stack_ordered_aligned_normalized}}
    \subfloat[\tiny\begin{tabular}{ll}
        type=PIE & order=CORRECT\_LAST \\
        normalize=TRUE & align=CORRECT\_CENTER
    \end{tabular}]{{\includegraphics[width=0.49\textwidth]{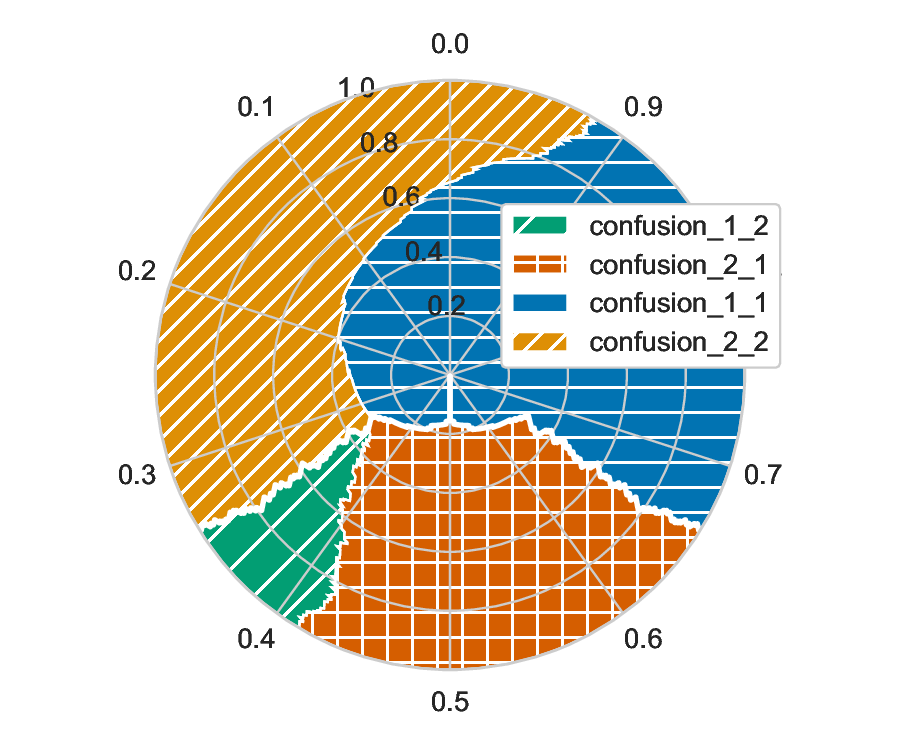}}\label{fig:example_2_classes_pie}}

    \caption{Further variants of SCORE for the 2-class setting. Ordering, alignment, and normalization are used to highlight different aspects of the confusion stack.}
    \label{fig:score_2_classes_variants}
\end{figure}

To improve the interpretability even more, SCORE offers options to order, align, and normalize the elements of the confusion stack in different ways to ease understanding (see the GitHub page for details). Figure~\ref{fig:score_2_classes_variants} shows some variants.
The stack in Fig.~\ref{fig:example_2_classes_stack_ordered} is ordered such that wrong confusions are at the bottom while correct confusions are on the top. This is especially useful when the border between wrong and correct confusions is aligned horizontally as in Fig.~\ref{fig:example_2_classes_stack_ordered_aligned}. Here, the positive y-axis corresponds to the count of correct samples, while the negative y-axis reflects the number of errors. When normalizing the counts with the number of accepted samples $|\mathcal{X}_{\theta}|$, as in Fig.~\ref{fig:example_2_classes_stack_ordered_aligned_normalized}, the positive and negative part of the y-axis are the accuracy and the error rate, respectively. This plot shows clearly, how certain the classifier is about \emph{confusion\_1\_1} as it strongly dominates the plot for lower acceptance rates. The normalized stack can be visualized as a pie chart (Fig.~\ref{fig:example_2_classes_pie}). The radius in such a plot corresponds to the acceptance rate, which decreases towards the center of the pie chart, while the angle reflects the stacked confusions. Here, the line with an angle of zero is aligned to the center of the correct confusions. The resulting symmetry of the plot makes it easier to access the share of correct and wrong classifications.

SCORE can be used for data with more than two classes. For this it offers an option to condense errors of each true class $c_{\text{t}}$ to a single confusion case.

\section{Conclusion}\label{sec:conclusion}
We argue that commonly used visualizations of reject options mask too much information and thus are hard to interpret even for experts.
Stacked Confusion Reject Plots (SCORE) display details about an experiment's class distribution and classifier behavior and hence are more intuitive. 
We showed the options of SCORE on artificial data and discussed how to reveal different aspects of a classifier with a reject option, e.\,g.\ in case of medical data.
The code is a public Python package, which we like to develop further with the community.

\pagebreak


\begin{footnotesize}

\bibliographystyle{unsrt}
\bibliography{main}

\end{footnotesize}


\end{document}